\documentclass{article}

    \PassOptionsToPackage{numbers, compress}{natbib}

    \usepackage[preprint]{neurips_2023}

\usepackage[utf8]{inputenc} 
\usepackage[T1]{fontenc}    
\usepackage{hyperref}       
\usepackage{url}            
\usepackage{booktabs}       
\usepackage{amsfonts}       
\usepackage{nicefrac}       
\usepackage{microtype}      
\usepackage{xcolor}         
\usepackage{bm}
\usepackage{bbm}

\usepackage{amsmath}
\usepackage{amsthm}
\usepackage{amssymb}
\usepackage{graphicx}
\usepackage{subcaption}
\usepackage{multirow}
\usepackage{float} 

\theoremstyle{plain}
\newtheorem{theorem}{Theorem}[section]
\newtheorem{proposition}[theorem]{Proposition}
\newtheorem{lemma}[theorem]{Lemma}
\newtheorem{corollary}[theorem]{Corollary}
\theoremstyle{definition}
\newtheorem{definition}[theorem]{Definition}
\newtheorem{assumption}[theorem]{Assumption}
\theoremstyle{remark}

\title{Towards the Sparseness of Projection Head in Self-Supervised Learning}

\author{
    Zeen Song$^{1,2}$\thanks{Equal Contribution},
    Xingzhe Su$^{1,2}$\footnotemark[1],
    Jingyao Wang$^{1,2}$\footnotemark[1],
    Wenwen Qiang$^2$\thanks{Corresponding author}, 
    Changwen Zheng$^2$, 
    Fuchun Sun$^3$\\
    University of Chinese Academy of Sciences$^1$,\\
    Institute of Software Chinese Academy of Sciences$^2$,\\
    Tsinghua University$^3$\\
    \small\texttt{\{songzeen22,suxingzhe18,wangjingyao22\}@mails.ucas.ac.cn}, \\
    \texttt{\{qiangwenwen,changwen\}@iscas.ac.cn}, \\ \texttt{fcsun@tsinghua.edu.cn}
}

\begin{document}

\maketitle

\begin{abstract}
  In recent years, self-supervised learning (SSL) has emerged as a promising approach for extracting valuable representations from unlabeled data. One successful SSL method is contrastive learning, which aims to bring positive examples closer while pushing negative examples apart. Many current contrastive learning approaches utilize a parameterized projection head. Through a combination of empirical analysis and theoretical investigation, we provide insights into the internal mechanisms of the projection head and its relationship with the phenomenon of dimensional collapse. Our findings demonstrate that the projection head enhances the quality of representations by performing contrastive loss in a projected subspace. Therefore, we propose an assumption that only a subset of features is necessary when minimizing the contrastive loss of a mini-batch of data. Theoretical analysis further suggests that a sparse projection head can enhance generalization, leading us to introduce SparseHead - a regularization term that effectively constrains the sparsity of the projection head, and can be seamlessly integrated with any self-supervised learning (SSL) approaches. Our experimental results validate the effectiveness of SparseHead, demonstrating its ability to improve the performance of existing contrastive methods.
\end{abstract}

\section{Introduction}
The objective of self-supervised learning (SSL) lies in the extraction of valuable representations from unlabeled data\cite{ericssonSelfSupervisedRepresentationLearning2022, bengio2013representation, liu2022self}. One prominent method that has achieved significant success in SSL is contrastive learning\cite{chenSimpleFrameworkContrastive2020, oordRepresentationLearningContrastive2019, heMomentumContrastUnsupervised2020}. This approach involves considering augmented samples as positive pairs, while others within the same batch serve as negative samples. The primary goal of contrastive learning is to minimize the distances between positive pairs, while simultaneously maximizing the distances between negative pairs.

One of the most famous framework of contrastive learning is SimCLR \cite{chenSimpleFrameworkContrastive2020}, which has achieved state-of-the-art performance in visual representation learning tasks. One particularly intriguing aspect of SimCLR is the introduction of a nonlinear projection head, which significantly enhances the quality of the learned representations. Empirical observations reveal that the quality of the features learned prior to the projection head, referred to as the \textit{representation}, surpasses that of the features obtained after the projection head, known as the \textit{embedding}, in downstream classification tasks (with an improvement of over 10\%) \cite[Section 4.2]{chenSimpleFrameworkContrastive2020}. 

Despite the effectiveness of the projection head, there is a limited understanding of its internal mechanism due to the lack of existing research on this topic. However, some recent works \cite{hua2021feature, jingUnderstandingDimensionalCollapse2022} argue that the poor performance of embeddings is a result of a phenomenon called dimensional collapse. This phenomenon occurs when the learned embeddings span a low-dimensional space, leading to a loss of information.

To gain a deeper understanding of the role of the projection head, we conducted a series of empirical experiments, which are presented in Section \ref{sec:motivation}. In these experiments, we visualized the log-scaled eigenvalues of the learned representations and embeddings obtained from different combinations of feature extractors and projection heads. Our findings are as follows:
i) The inclusion of a projection head, whether non-linear or linear, leads to significantly larger eigenvalues in the learned representations compared to representations trained without a projection head.
ii) When a projection head is incorporated, the eigenvalues of the embeddings are noticeably lower than those of the representations.
iii) Even with the presence of a projection head, in cases where the representation space is larger, the dimensional collapse phenomenon cannot be entirely mitigated.

Building upon the aforementioned findings, this paper addresses the core assumption that minimizing the contrastive loss of a mini-batch of samples can be achieved using only a subset of features from the representation space. This assumption is visually illustrated in Figure \ref{fig:sparse task}. To further explore the implications of this assumption, we conduct a theoretical analysis in Section \ref{sec:Theo Anal}, where we investigate the relationship between the sparsity of the projection head, the generalization ability, and the discriminative performance. This analysis serves as a motivation for our proposed method called SparseHead, which is a regularization term designed to effectively constrain the sparsity of the projection head. In Section \ref{sec:Exp}, we evaluate the performance of SparseHead in various downstream tasks, aiming to enhance the quality of learned representations.


Our contributions can be summarized as follows:
\begin{itemize}
    \item We empirically demonstrate the significant impact of incorporating a projection head in contrastive learning. This leads to improved quality of learned representations compared to representations trained without a projection head. Additionally, we highlight the effectiveness of projecting the representation into a lower-dimensional space. However, we also identify the limitations of simply adding a projection head.
    \item We undertake a theoretical analysis to explore the impact of the sparsity of the projection head. Our investigation provides evidence supporting the claim that a sparse projection head can improve the generalization capabilities of learned representations. Additionally, we validate the correctness of the key assumption by demonstrating the discriminative ability of the learned representations in lower-dimensional spaces. Notably, we establish a lower bound for the discriminative ability, which is determined by the sparsity constraint hyperparameter.
    \item We introduce SparseHead, a regularization term designed to effectively constrain the sparsity of the projection head. SparseHead can be seamlessly integrated with any self-supervised learning (SSL) approaches. Through a series of experiments, we validate the effectiveness of SparseHead and demonstrate its ability to enhance both generalization and discriminative performance in learned representations.
\end{itemize}


\section{Related Works}
\textbf{Contrastive Learning in Practice.} Contrastive learning emerged as a powerful technique in the field of unsupervised representation learning, initially introduced from a mutual information perspective \cite{gutmannNoisecontrastiveEstimationNew2010, chenExploringSimpleSiamese2021, heMomentumContrastUnsupervised2020, koRevisitingContrastiveLearning2022, oordRepresentationLearningContrastive2019}. Notably, SimCLR \cite{chenSimpleFrameworkContrastive2020} made significant strides in narrow the performance gap between unsupervised and supervised methods. However, when used in practical application, the performance of SimCLR is hindered by the need for a large number of negative samples. To relax the need of numerous negative samples, negative-related techniques, such as BYOL \cite{grillBootstrapYourOwn2020}, SimSiam \cite{chenExploringSimpleSiamese2021}, MoCHi\cite{kalantidis2020hard}, PBNS\cite{ge2021robust}, and DCL\cite{chuang2020debiased} have been introduced, which utilize stop-gradient and asymmetric predictor approaches, respectively. Another set of solutions aims to overcome the trivial solution problem. Barlow Twins \cite{zbontarBarlowTwinsSelfSupervised2021}, SSL-HSIC \cite{li2021self}, VICReg \cite{bardesVICRegVarianceInvarianceCovarianceRegularization2022}, W-MSE \cite{ermolovWhiteningSelfSupervisedRepresentation2021}, and CW-RGP \cite{wengInvestigationWhiteningLoss2022} suggest constraining the covariance matrix of the sample features to be a unit matrix. Additionally, the augmentation technique in contrastive learning has also been explored. MetAug \cite{li2022metaug} proposes a learnable augmentation method for contrastive learning. AugSelf\cite{lee2021improving} proposes a augmentaion-aware approach. In our work, we propose a novel sparsity constraint on the projection head to enhance generalization, distinguishing our approach from previous works in this field.

\textbf{Theoretical Findings.} Contrastive learning has not only led to methodological advancements but has also provided significant theoretical insights into its underlying principles \cite{zhangHowDoesSimSiam2022, wenMechanismPredictionHead2022, balestrieroContrastiveNonContrastiveSelfSupervised2022, caoEmpiricalStudyDisentanglement2022}. The pioneers of theoretical guarantees for contrastive loss were Saunshi et al. \cite{saunshiTheoreticalAnalysisContrastive2019}, who explored the impact of negative samples on contrastive learning performance. Ash et al. \cite{ashInvestigatingRoleNegatives2021} delved into the role of negatives and introduced the "collision-coverage" problem within the same framework. However, the work of Awasthi et al. \cite{awasthiMoreNegativeSamples2022} demonstrated that, under certain circumstances, increasing the number of negative samples does not necessarily impair performance. Understanding the optimization objectives of contrastive learning, Wang et al. \cite{wangUnderstandingContrastiveRepresentation2020} established that the contrastive loss aims to optimize both alignment and uniformity. Furthermore, the phenomenon of dimensional collapse in contrastive learning has been analyzed by DirectCLR \cite{jingUnderstandingDimensionalCollapse2022} and MetaMask \cite{li2022metamask}. The augmentation graph and graph connectivity have also been proposed as theoretical analyses \cite{haochenProvableGuaranteesSelfSupervised2021, wangChaosLadderNew2022, balestrieroContrastiveNonContrastiveSelfSupervised2022}. Despite these contributions, Saunshi et al. \cite{saunshiUnderstandingContrastiveLearning2022} argue that the augmentation of overlap alone is insufficient to explain the success of contrastive learning. Exploring the causal effects of contrastive learning, SSL-PICS \cite{von2021self} and ICL-MSR \cite{qiangInterventionalContrastiveLearning2022} have investigated the causal effects and developed structural causal models to enhance performance. Garrido et al. \cite{garrido2022duality} examined the duality between contrastive and non-contrastive self-supervised learning. Additionally, Wen et al. \cite{wenMechanismPredictionHead2022} have focused on understanding the mechanism of the prediction head in non-contrastive self-supervised learning. In this paper, we contribute to the theoretical analysis by providing a detailed examination of the properties of the projection head in contrastive learning.

\begin{figure}[H]
    \centering
     \begin{subfigure}{0.45\textwidth}
         \centering
         \includegraphics[width=\textwidth]{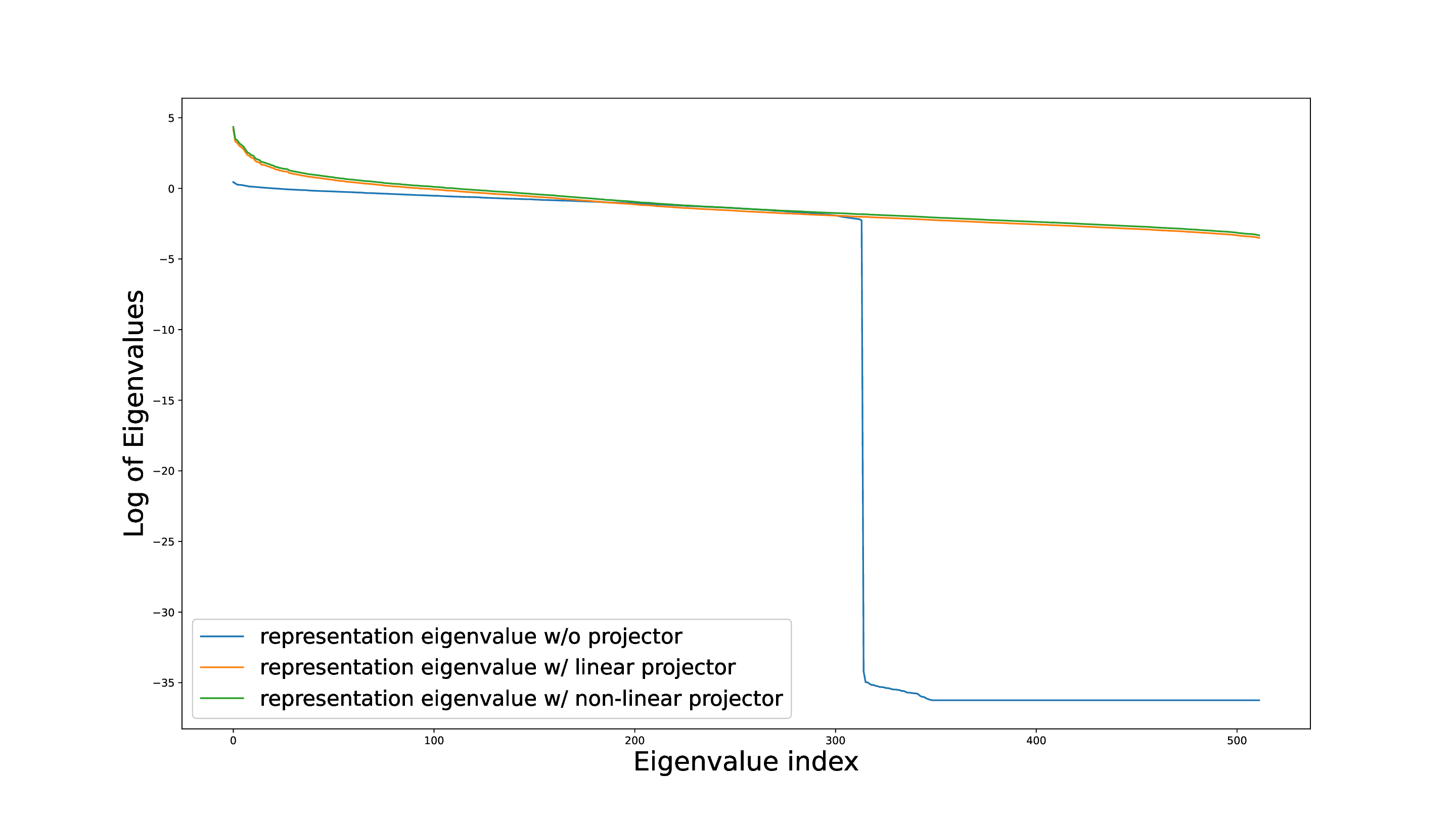}
         \caption{$\sigma_r$ with $h_\phi$ is linear, non-linear and identity}
         \label{fig:1.1}
     \end{subfigure}
     \begin{subfigure}{0.45\textwidth}
         \centering
         \includegraphics[width=\textwidth]{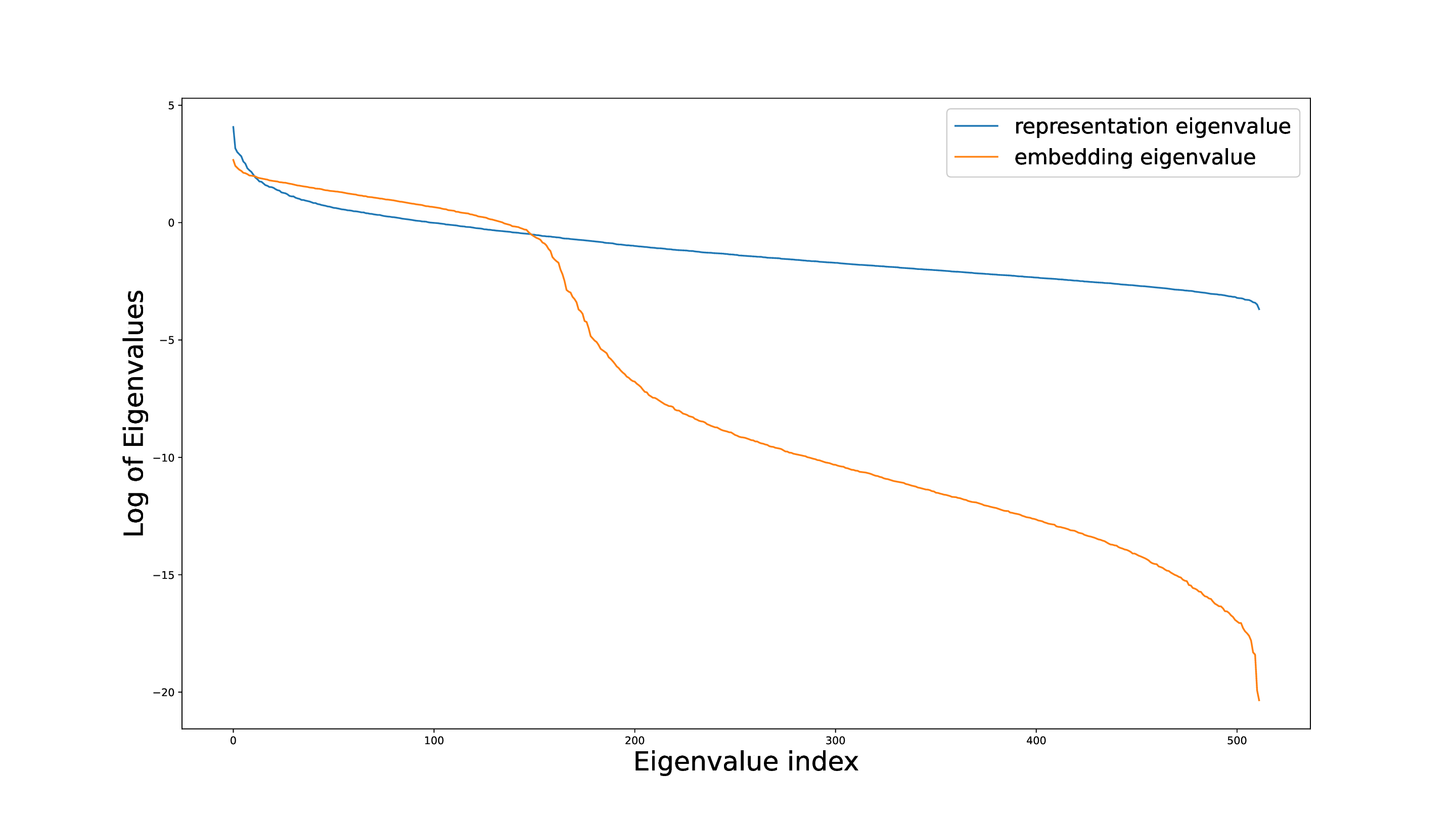}
         \caption{$\sigma_r$ and $\sigma_z$ with $h_\phi$ is linear}
         \label{fig:1.2}
     \end{subfigure}
     \begin{subfigure}{0.45\textwidth}
         \centering
         \includegraphics[width=\textwidth]{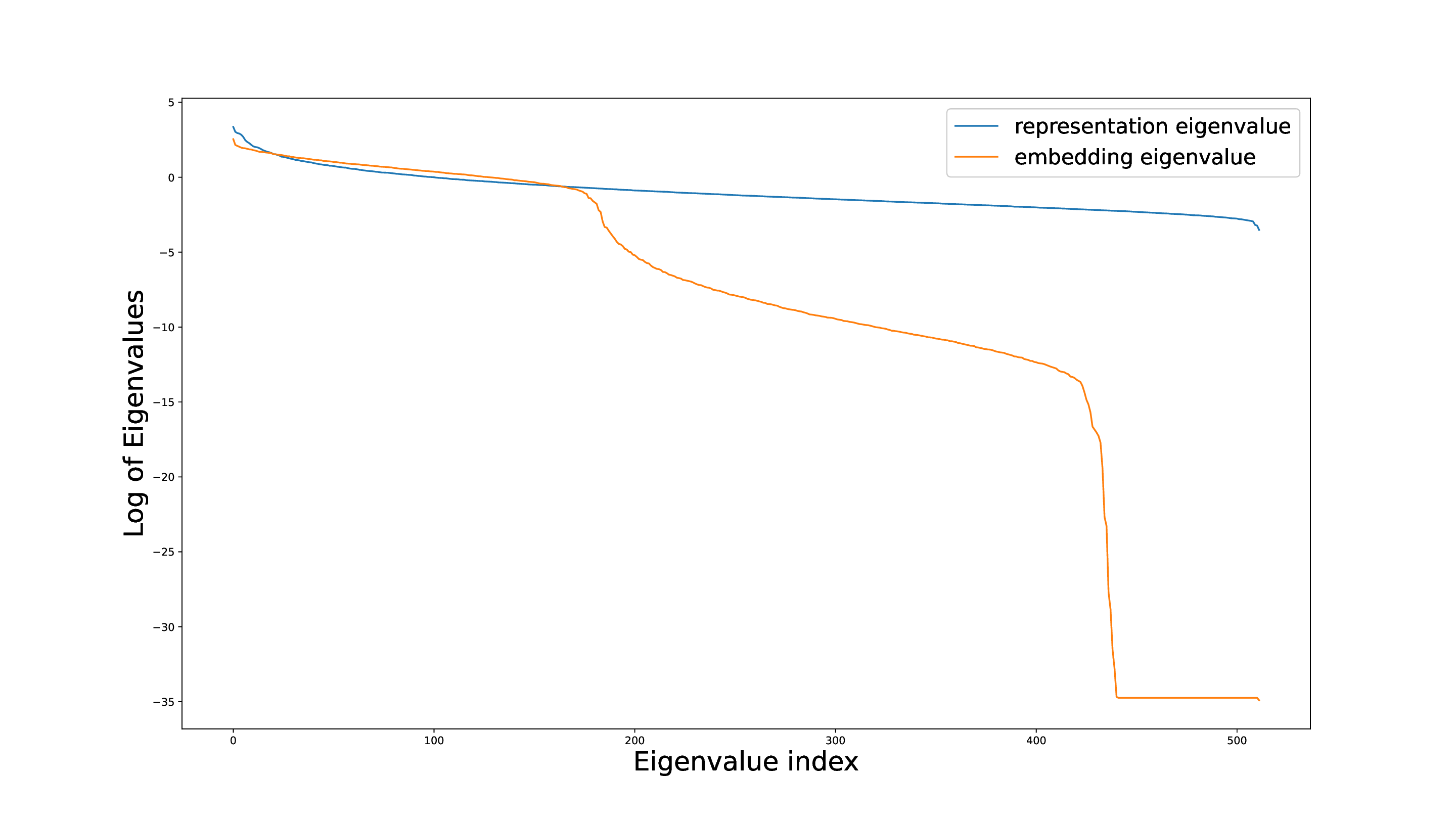}
         \caption{$\sigma_r$ and $\sigma_z$ with $h_\phi$ is non-linear}
         \label{fig:1.3}
     \end{subfigure}
     \begin{subfigure}{0.45\textwidth}
         \centering
         \includegraphics[width=\textwidth]{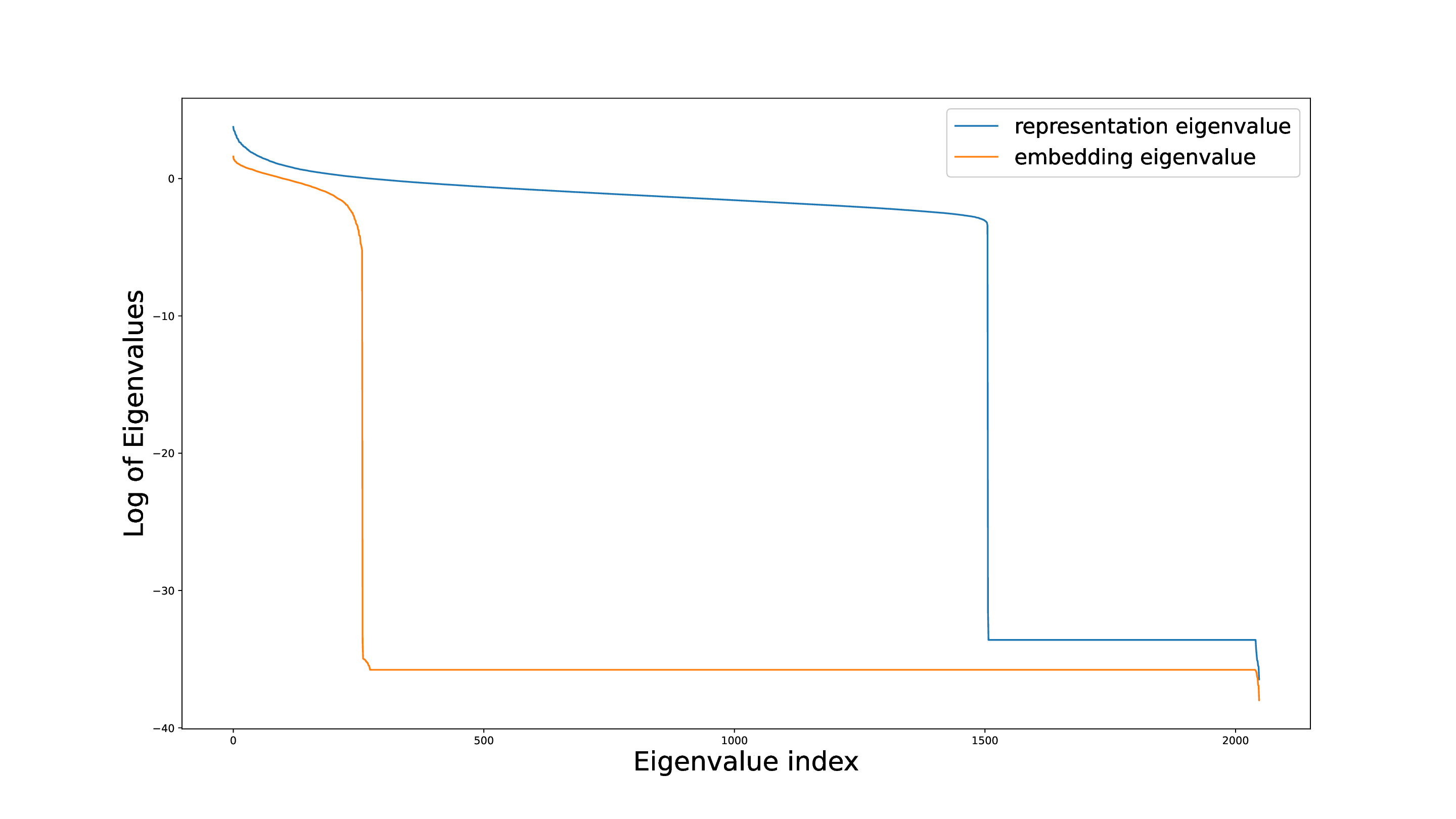}
         \caption{$\sigma_r$ and $\sigma_z$ with $f_\theta$ is ResNet-50}
         \label{fig:1.4}
     \end{subfigure}
        \caption{The eigenvalue spectras of different representations and embeddings}
        \label{fig:toy experiment}
\end{figure}

\section{Preliminary Study}
Given a dataset $\mathcal{D} = \{\boldsymbol{x}\}$ and a  data augmentations distribution $\mathcal{A}$, the learning objective of contrastive learning can be written as:
\begin{align}\label{eq:objective_CL}
    \min_{\theta,\phi}\mathcal{L}_{\text{contrastive}}(\theta,\phi;\mathcal{D},\mathcal{A}) &:=\mathbb{E}_{\mathcal{B}_{\boldsymbol{x}} \sim \mathcal{D}}[\ell(\theta,\phi;\mathcal{Z}_{\boldsymbol{x}})],\\\nonumber
    s.t. \quad \mathcal{Z}_{\boldsymbol{x}} &= \left\{h_\phi(f_\theta(a^l(\boldsymbol{x})))\mid \boldsymbol{x}\in\mathcal{B}_{\boldsymbol{x}},a^l\sim\mathcal{A}, l \in\{1, 2\}\right\}\ .
\end{align}

In Equation \ref{eq:objective_CL}, $\mathcal{B}_{\boldsymbol{x}}$ is a mini-batch sampled from $\mathcal{D}$,  $f_\theta$ denotes the feature extractor and $h_\phi$ is the projection head. $\ell$ is the InfoNCE loss of this mini-batch of samples:

\begin{equation}\label{eq:infonce}
\ell(\theta,\phi;\mathcal{Z}_{\boldsymbol{x}}) = -\sum_{\boldsymbol{z}^l\in\mathcal{Z}_{\boldsymbol{x}}}\log(\frac{\exp(\cos(\boldsymbol{z}^l, \boldsymbol{z}^{3-l})/\tau)}{\sum_{\boldsymbol{z}\in\mathcal{Z}_{\boldsymbol{x}} \setminus \{\boldsymbol{z}^l\}}\exp(\cos(\boldsymbol{z}^l, \boldsymbol{z})/\tau)})
\end{equation}

Where $\cos(\boldsymbol{u},\boldsymbol{v}) = \boldsymbol{u}^T\boldsymbol{v}/\Vert\boldsymbol{u}\Vert \Vert\boldsymbol{v}\Vert$ and $\tau$ is a hyperparameter for temperature scaling. 

The feature extractor $f_\theta$ is parameterized as a backbone neural network, and the projection head $h_\phi$ is implemented as a multi-layer MLP or a single-layer weight matrix. We denote the extracted \textbf{Representation} as $\boldsymbol{r} = f_\theta(\boldsymbol{x}) \in \mathbb{R}^d$. And the projected \textbf{Embeddings} as $\boldsymbol{z} = h_\phi(\boldsymbol{r}) \in \mathbb{R}^m$. Equation \ref{eq:infonce} implies that the similarity between $\boldsymbol{z}^l$ and $\boldsymbol{z}^{3 - l}$ is maximized while the similarities between $\boldsymbol{z}^l$ and negative samples $\mathcal{Z}_{\boldsymbol{x}} \setminus \{\boldsymbol{z}^l\}$ are minimized.

After the training phase, the projection head $h_\phi$ is discarded. The trainable weights in feature extractor $f_\theta$ is frozen and a coefficient matrix $\bar{W}$ is trained on the training set $\mathcal{D}_\text{train}=\{(\boldsymbol{x},y)\}$ of downstream task, the label $y$ can be either class indices or realvalues, depending on the property of tasks. The performance of $f_\theta$ on the downstream task can be evaluated with maximum likelihood estimation (MLE) on the test set $\mathcal{D}_\text{test}=\{(\boldsymbol{x},y)\}$ as below:

\begin{align}\label{eq:downstream}
    \mathcal{L}_{\text{downstream}}(\Bar{W};f_\theta, \mathcal{D}_{\text{test}}, \mathcal{D}_{\text{train}}) &= \mathbb{E}_{(\boldsymbol{x},y) \sim \mathcal{D}_{\text{test}}}[-\log P(y ; \bar{W} f_{\theta}(\boldsymbol{x}))], \\
    s.t. \quad \bar{W} &= \arg\max_{\Bar{W}} \mathbb{E}_{(\boldsymbol{x},y) \sim \mathcal{D}_{\text{train}}}[\log P(y ; \Bar{W} f_{\theta}(\boldsymbol{x}))] \notag
\end{align}

\begin{figure}[t]
    \centering
    \includegraphics[width=\textwidth]{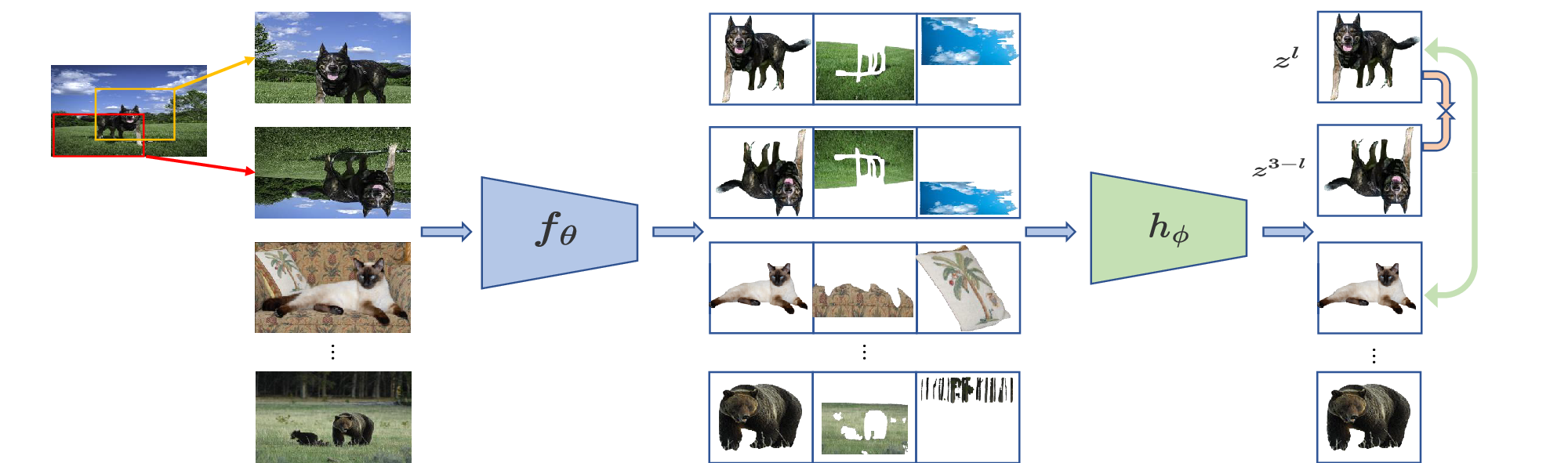}
    \caption{The illustration of our assumption that only a subset of features are used when minimizing contrastive loss. The representation vector comprises subject features, foreground features, and background features extracted from the images. Remarkably, we observe that minimizing the contrastive loss solely necessitates the use of the subject feature from these images.  }
    \label{fig:sparse task}
\end{figure}

\section{Empirical Analyses}
\label{sec:motivation}


To gain an intuitive understanding of the role of the projection head in contrastive learning, we conducted a series of experiments. We utilized the SimCLR method \cite{chenSimpleFrameworkContrastive2020} on the STL-10 dataset \cite{coatesAnalysisSingleLayerNetworks2011} to train the feature extractor $f_\theta$ with various projection heads $h_\phi$. Three types of projection heads were explored: a fixed identity matrix $h_\phi = I$, a single linear layer $h_\phi = W$, and a multi-layer non-linear fully-connected layer. The architecture of the feature extractor $f_\theta$ employed ResNet-18 and ResNet-50 \cite{heDeepResidualLearning2016}. The dimensions of the learned representations and embeddings were set as $d = m = 512$ for ResNet-18 and $d = m = 2048$ for ResNet-50. Next, we collected all the embedding and representation vectors from the test set $\boldsymbol{X}_{\text{test}}$, denoted as $\boldsymbol{R}$ and $\boldsymbol{Z}$, respectively. Covariance matrices $C_r \in \mathbb{R}^{d \times d}$ and $C_z \in \mathbb{R}^{m \times m}$ were computed for the representation and embedding vectors, respectively. To gain insights into the structures of these covariance matrices, we performed Eigenvalue Decomposition (EVD) on them, resulting in eigenvalues $\sigma_r$ and $\sigma_z$ for $C_r$ and $C_z$. Figure \ref{fig:toy experiment} illustrates the log-scaled eigenvalue spectra of $C_r$ and $C_z$ under different $h_\phi$ and $f_\theta$.



Firstly, as depicted in Figure \ref{fig:1.1}, we observe the impact of different projection heads on the rank of $C_r$. When no projection head is used ($h_\phi=I$), numerous eigenvalues are zero, indicating that the rank of $C_r$ is less than $d$. This observation suggests the occurrence of dimensional collapse. Secondly, Figures \ref{fig:1.2} and \ref{fig:1.3} demonstrate that when a linear or nonlinear projection head is employed, the rank of $C_r$ is equal to $d$, thereby avoiding dimensional collapse. However, we note that the rank of $C_z$ is lower than the rank of $C_r$ in these cases, implying a loss of information in the embedding space. Lastly, Figure \ref{fig:1.4} highlights that even with the presence of a projection head, dimensional collapse can still occur in larger networks and representation spaces. Observations are summarized as follows: 1) The representation vectors $\boldsymbol{R}$ contain more informative features compared to the embedding vectors $\boldsymbol{Z}$; 2) Performing contrastive learning on the embedding space can yield improvements in representation learning; 3) Despite the presence of a projection head, the issue of dimensional collapse cannot be completely mitigated.

These observations provide the motivation to propose the core assumption of this paper, which asserts that minimizing the contrastive loss of a mini-batch of samples can be achieved by using only a subset of features from the representation space.

We illustrate this idea in Figure \ref{fig:sparse task}, where a dog image serves as the original sample and is cropped twice to obtain the anchor and positive sample. The negative samples consist of images of other animals. Remarkably, in this scenario, the features extracted from the animal subject in the original image are adequate for discerning between the positive and negative samples. In the subsequent subsection, we formalize this hypothesis within the framework of multi-task learning.


\section{Theoretical Analyses}
\label{sec:Theo Anal}
\subsection{Bridging contrastive learning and unsupervised meta-learning}
\label{subsec:4.1}
Multi-task learning (MTL) is a subfield of machine learning that aims to enhance the performance and generalization of a learning system by simultaneously training it on multiple related tasks\cite{ruder2017overview,caruana1997multitask}. The fundamental concept behind MTL is to learn a task-agnostic feature extractor capable of capturing relevant features for each task from a shared input space, enabling the learning system to handle multiple tasks effectively\cite{zhang2018overview}. Meta-learning builds upon MTL by enabling the model to rapidly adapt to new tasks that were not encountered during training\cite{finnModelagnosticMetalearningFast2017, wang2021bridging}. It focuses on developing models that can learn from a distribution of tasks and generalize to new, unseen tasks efficiently. Contrastive learning (CL) shares a similar objective with MTL and meta-learning, aiming to learn transferable representations in a task-agnostic manner from unlabeled data. CL can be viewed as the unsupervised counterpart of meta-learning, and its learning objective is closely related to metric-based meta-learning \cite{niCloseRelationshipContrastive2022,jangUnsupervisedMetalearningFewshot2023, leeSelfSupervisedSetRepresentation2023}.

If we consider the anchor $\boldsymbol{z}^l$ as a query sample and $\mathcal{Z}_{\boldsymbol{x}}$ as the support set, Equation \ref{eq:infonce} can be seen as equivalent to an $N$-way 1-shot few-shot classification task. In this case, we can focus on the linear head $h_\phi=W$, and introduce the indicator variable $y_{\mathbbm{1}}=\mathbbm{I}(\boldsymbol{z}^l, \boldsymbol{z}^{3-l})$ to denote whether $\boldsymbol{z}^l$ and $\boldsymbol{z}^{3-l}$ form a positive pair. With these considerations, Equation \ref{eq:infonce} can be alternatively expressed as follows:
\begin{equation}\label{eq:mle_contrastive}
    \ell(\theta,\phi;\mathcal{Z}_x) = -\sum_{\boldsymbol{z}^l\in\mathcal{Z}_x}\log P(y_{\mathbbm{1}};W f_\theta(a^l(\boldsymbol{x})))
\end{equation}
Combining Equation \ref{eq:mle_contrastive} and Equation \ref{eq:downstream}, we can conclude that the goal of contrastive learning is to learn a task-agnostic feature extractor $f_{\theta^*}$ that apply to all tasks. 
Hence, we define the task $t$ as a tuple $t = (\mathcal{B}_{\boldsymbol{x}}, \mathcal{A}_{\mathcal{B}_{\boldsymbol{x}}})$ sampled from the distribution $P(t)$, where $\mathcal{A}_{\mathcal{B}_{\boldsymbol{x}}}$ represents the i.i.d. augmentation for mini-batch $\mathcal{B}_{\boldsymbol{x}}$. 

\subsection{Main Results}
\label{sec:main results}

We begin by assuming the existence of a ground-truth (GT) feature extractor $f_{\theta^*}$, which generates the ground-truth representation $f_{\theta^*}(\boldsymbol{x})$ for any given sample $\boldsymbol{x}\in \mathbb{R}^X$. The objective of multi-task learning is to learn a feature extractor $f_{\hat{\theta}}$ that approximates the GT feature extractor. To formalize this idea, we introduce the concept of the ground-truth equivalent representation (GTE) as follows:

\begin{definition}
\label{def:1}
(Ground-truth equivalent representation) A learned encoder function $f_{\hat{\theta}}:\mathbb{R}^X \rightarrow \mathbb{R}^d$ is said to be equivalent w.r.t the ground-truth representation $f_{\theta^*}$ when there exists an diagonal matrix $\Lambda$ and a permutation matrix $P$ such that, for all $\boldsymbol{x} \in \mathcal{X}, f_{\hat{\theta}} = \Lambda P f_{\theta^*}(\boldsymbol{x})$.
\end{definition}

Roeder et al. \cite{roeder2021linear} demonstrated that contrastive learning can achieve linear identifiability for any feature extractor $f_{\theta}$ with parameters $\theta \in \Theta$. In other words, for any $x \in \mathcal{X}$, there exists an invertible matrix $L$ such that $f_{\hat{\theta}}(\boldsymbol{x}) = Lf_{\theta^*}(\boldsymbol{x})$, where $\{\theta^*, \hat{\theta}\} \subseteq \Theta$ and $f_{\theta^*}$ represents the optimal feature extractor. According to Definition \ref{def:1}, this implies that the ground-truth equivalent (GTE) representation $f_{\hat{\theta}}$ is a specific instance of linear identifiability, with $L = \Lambda P$. We can easily prove that the GTE representation does not adversely affect generalization, as defined in Definition \ref{def:1}, by considering the following:

\begin{proposition}
\label{prop:1}
(Generalization via GTE representation) 
\begin{align}
    \hat{\theta} &= \mathop{\arg\min}_{\Tilde{\theta}} \mathbb{E}_{t \sim P(t)} \mathbb{E}_{x\in\mathcal{B}^t_x} -\log P(y_{\mathbbm{1}};W^tf_{\Tilde{\theta}}(\boldsymbol{x})) \\
     W^t &= -\arg\min_{\Tilde{W}} \mathbb{E}_{x\in \mathcal{B}_x} \log P(y_{\mathbbm{1}};\Tilde{W}f_{\Tilde{\theta}}(\boldsymbol{x}))
\end{align}
    where $W^t$ is the task-specific ground-truth classifier and $\forall x \in \mathcal{X}, f_{\hat{\theta}} = \Lambda P f_{\theta^*}(x)$.
\end{proposition}
The detailed proof is illustrated in Appendix. Based on the previous section, we concluded that only a subset of features is used to compare losses for all tasks. We can thus make the following hypothesis about the $L_{2,0}$ norm of task-specific ground-truth classifiers $W^t$:
\begin{assumption}
    \label{ass:1}
    (Task specific classifier)
    for a given task $t$ and a learned feature extractor $f_{\hat{\theta}}$, the $L_{2,0}$ norm of GT projection head $\Vert W^t \Vert_{2,0} < d$, indicating that the task-specific head is sparse, and only a subset of features are used for task $t$.
\end{assumption}
Let $S^t$ be the support of the matrix $W^t$, i.e. the set of features that are relevant for predicting $y$ in the $t$-th task: $S^t := \{j\in D | \Vert W^t_{:j} \Vert_{2,0} \not= 0\}$, where $D:=\{i\}_{i=1}^d$. Note that $S^t$ is unknown to the learner. We assume that $W$ follows a distribution $P(W)$, and the support $S$ follows a distribution $P(S)$. We can write $P(W)$ as a mixture of conditional distributions: $P(W) = \sum_{S\in\mathcal{P}(D)}P(S)P(W|S)$. Here, $\mathcal{P}(D) = \{S|S\subseteq D\}$ is the power set of $D$. We also define $\mathcal{S} := \{S\in \mathcal{P}(D) | P(S) > 0\}$ as the set of non-trivial supports. The support $S$ and $W$ satisfy the following assumptions.
\begin{assumption}
    \label{ass:2}
    (Intra-support sufficient variance) $\forall S \in \mathcal{S}$ and all $\boldsymbol{u} \in \mathbb{R}^{|S|} \setminus \{0\}$. $\mathbb{E}_{P(W)} [W_{:S} \boldsymbol{u} = 0 | S] = 0$.
\end{assumption}
Assumption \ref{ass:2} indicates that the distribution $P(W|S)$ does not lie within a specific subspace, thereby indicating that all support $S\in \mathcal{S}$ are non-trivial.
\begin{assumption}
    \label{ass:3}
    (Non-trivial Feature) $\forall j \in D$, $\bigcup_{S\in \mathcal{S}|j\not\in S}S=D\setminus \{j\}$.
\end{assumption}
Assumption \ref{ass:3} asserts that for every feature $j$, there exists a collection of supports $S$ such that the union of these supports includes all features except for feature $j$ itself. This implies that the union of supports is sufficiently diverse to cover all features, and no feature appears in all tasks. Building upon Assumptions \ref{ass:2} and \ref{ass:3}, we can establish a proof that the following bi-level optimization problem guarantees the achievement of the ground-truth equivalent (GTE) representation.
\begin{theorem}
\label{theo:1}
(GTE via sparsity) Let $\hat{\theta}$ be a minimizer of 
\begin{align}
     &\min \mathbb{E}_{t \sim P(t)} \mathbb{E}_{x\in\mathcal{B}^t_x} -\log P(y_{\mathbbm{1}};\hat{W}^tf_{\Tilde{\theta}}(\boldsymbol{x})) \\
     & s.t. \quad \forall W \in \mathcal{W}, \hat{W}^t \in \mathop{\arg\min}_{\substack{\Tilde{W} s.t. \\ \Vert \Tilde{W} \Vert_{2,0} \leq \Vert W \Vert_{2,0}}} \mathbb{E}_{x\in \mathcal{B}_x} -\log P(y;\Tilde{W}f_{\hat{\theta}}(x))
\end{align}
If $f_{\hat{\theta}}$ is continuous for all $\hat{\theta}$, $f_{\hat{\theta}}$ is equivalent to ground-truth feature extractor  $f_{\theta^*}$.
\end{theorem}
The detailed proof is provided in Appendix. This section presents theoretical analyses to showcase that imposing sparsity constraints on the projection head can encourage the feature extractor to learn ground-truth equivalent (GTE) representations, which in turn enhance generalization performance.

\subsection{Discriminant Analyses}
In order to provide further insight into the justification for Assumption \ref{ass:1}, we delve deeper into its rationale. Figure \ref{fig:sparse task} visually illustrates the intuition behind this assumption. In this subsection, we specifically explore the discriminative capability of low-rank features.
As we have seen in the previous section, contrastive learning involves treating each batch as a classification problem, where the objective is to make the embeddings of the same class closer and the embeddings of different classes farther apart. Thus, the discriminability of the embeddings plays a crucial role in representation learning.
 It has been proven in \cite{beyer1999nearest} that as dimensionality increases, the distance to the nearest data point approaches the distance to the farthest data point. To quantify the discriminativeness of the feature extractor, we can utilize the \textit{min-max distance ratio}, which is defined as:
\begin{equation}
\label{eq:mmdr}
    M_{\theta, \phi}(d) = \frac{\delta^{\max}(d) - \delta^{\min}(d)}{\delta^{\min}(d)}
\end{equation}
Here, $\delta^{\max}(d) = \max\{\|\boldsymbol{z} - \boldsymbol{z}_i\|_2 \mid i = 1, \dots, n\}$ and $\delta^{\min}(d) = \min\{\|\boldsymbol{z} - \boldsymbol{z}_i\|_2 \mid i = 1, \dots, n\}$ represent the maximum and minimum distances between the embedding vector $\boldsymbol{z}$ and the other samples, respectively. When the dimensionality of the representation approaches infinity, \cite{beyer1999nearest} propose the following conclusion.
\begin{lemma}
\label{lem:1}
For any given embeddings $\boldsymbol{z},\boldsymbol{z}_1,\boldsymbol{z}_2,\dots,\boldsymbol{z}_n\in \mathbb{R}^d$  generated by feature extractor $f_\theta$ and $h_\phi$ of i.i.d. random data points $\boldsymbol{x}_i\in \mathcal{X}$, we have that 
\begin{equation}
    P(\lim_{d\rightarrow \infty}M_{\theta, \phi}(d)=0)=1
\end{equation}
\end{lemma}
According to Lemma \ref{lem:1}, as the dimensionality increases infinitely, the similarity between pairwise instances becomes less effective in providing contrast for distinguishing positive and negative pairs. In such cases, it is reasonable for the learning objective to prioritize low-dimensional embeddings over high-dimensional ones.
From Section \ref{sec:main results}, we can conclude that the incorporation of a sparse projection head has the potential to improve the generalization capabilities of representations by projecting them into a low-dimensional subspace. This is achieved through the inclusion of the regularization term $\lambda \|\tilde{W}\|_{2,1}$ in the optimization problem \ref{eq:problem}, which effectively limits the dimensionality of the embedding $z$. Expanding on this observation, we present a proof to demonstrate that the discriminative measurement $M_{\theta, \phi}(d)$ is lower bounded by the regularization parameter $\lambda$ in the following manner:
\begin{theorem}
\label{theo:2}
    (lower bounded min-max distance ratio)
    For any given embeddings $\boldsymbol{z},\boldsymbol{z}_1,\boldsymbol{z}_2,\dots,\boldsymbol{z}_n\in \mathbb{R}^d$  generated by learned feature extractor $f_{\hat{\theta}}$ and $h_{\hat{\phi}}$ of i.i.d. random data points $\boldsymbol{x}_i\in \mathcal{X}$, we have that 
    \begin{equation}
        P(\lim_{d\rightarrow \infty}M_{\hat{\theta}, \hat{\phi}}(d)>\lambda C(\mathcal{X}))=1
    \end{equation}
    where $M_{\hat{\theta}, \hat{\phi}}$ is the min-max distance ratio from the learned embedding $\boldsymbol{z}$ with Equation \ref{eq:problem} and $C(\mathcal{X})$ is a constant only related to original samples.
\end{theorem}
The detailed proof is provided in the Appendix. Theorem \ref{theo:2} implies that the learned embedding can successfully capture the intrinsic similarity during the minimization of the learning objective, thereby resulting in improved performance on downstream tasks.

\section{Method}
Theorem \ref{theo:1} establishes the link between sparsity and the GTE representations. However, the intractable nature of the $L_{2,0}$ norm motivates us to employ a relaxation technique. Specifically, we adopt the least absolute shrinkage and selection operator (lasso) \cite{tibshiraniRegressionShrinkageSelection1996, flexederGeneralizedLassoRegularization2010}, which serves as an approximation to the $L_{2,0}$ norm. By doing so, we arrive at the following corollary.
\begin{corollary}
    (Optimization with lasso) The Theorem \ref{theo:1} can be relaxed as the following optimization problem with a $L_{2,1}$ regularization term.
    \begin{align}
    \label{eq:problem}
        &\min_{\hat{\theta}} -\frac{1}{NT}\sum_{t=1}^T\sum_{x\in \mathcal{B}_x^t} \log P(y_{\mathbbm{1}};\hat{W}^t f_{\hat{\theta}}(a^l(x))) \\
        & s.t.\quad \forall t\in [T], \hat{W}^t \in \mathop{\arg\min}_{\Tilde{W}} -\frac{1}{N} \sum_{x \in \mathcal{B}_x^t} \log P(y_{\mathbbm{1}};\Tilde{W} f_{\hat{\theta}}(a^l(x))) + \lambda \Vert \Tilde{W} \Vert_{2,1}
    \end{align}
\end{corollary}

We therefore propose a regularization method called SparseHead. Specifically, we introduce a regularization term that encourages sparsity in the projection head. If the projection head is linear, represented as $h_\phi=W$, we apply the $L_{2,1}$ norm as the regularization term, given by $\lambda \Vert W \Vert_{2,1}$. On the other hand, if the projection head is non-linear, we use the $L_{2,1}$ norm of the weight matrix in the last layer as the regularization term. In the upcoming section, we present a series of experiments to demonstrate the effectiveness of the proposed SparseHead regularization term. These experiments provide insights into how sparsity promotion can improve the performance of self-supervised learning.

\begin{table}[t]
    \centering
    \caption{Classification accuracy for small, medium, and large datasets. The backbone is ResNet-18 for the first four datasets and ResNet-50 for the last two datasets.}
    \vspace{0.1cm}
    \resizebox{\linewidth}{!}{
    \begin{tabular}{l|cc|cc|cc|cc|cc|cc}
    \toprule
    \multirow{2}*{Methods} & \multicolumn{2}{c|}{CIFAR-10} & \multicolumn{2}{c|}{CIFAR-100} & \multicolumn{2}{c|}{STL-10} & \multicolumn{2}{c|}{Tiny ImageNet} & \multicolumn{2}{c|}{ImageNet-100} & \multicolumn{2}{c}{ImageNet}
    \\
    ~ & linear & 5-nn & linear & 5-nn & linear & 5-nn & linear & 5-nn & top-1 & top-5 & top-1 & top-5\\
    \midrule
    SimCLR \cite{chenSimpleFrameworkContrastive2020} &  91.80  & 88.42 & 66.83  & 56.57 & 90.51  & 85.68 & 48.82  & 32.86 &  70.15  & 89.75 & 69.32  & 89.15\\
    BYOL \cite{grillBootstrapYourOwn2020} &  91.73  & 89.26 & 66.60  & 56.82 & 91.86  & 88.61 & 51.01  & 36.14 &  74.89 & 92.83 & 74.31  & 91.62\\
    BarlowTwins \cite{zbontarBarlowTwinsSelfSupervised2021} &  90.88  & 88.78 & 66.67  & 56.39 & 90.71  & 86.31 & 49.74  & 33.61 & 72.88 & 90.99 & 73.22 & 91.01\\
    SimSiam \cite{chenExploringSimpleSiamese2021} &  91.51  & 89.31 & 66.73  & \textbf{56.87} & 91.92  & 88.54 & 50.92  & 35.98 & 74.78 & 92.84 & 71.33  & -\\
    W-MSE \cite{ermolovWhiteningSelfSupervisedRepresentation2021} &  91.99  & 89.87 & 67.64  & 56.45 & 91.65  & 88.49 & 49.22  & 35.44 & 75.33 & 92.78 & 72.56 & -\\
    SwAV \cite{caronUnsupervisedLearningVisual2020} &  90.17  & 86.45 & 65.23  & 54.77 & 89.12  & 84.12 & 47.13  & 31.07 &  75.77  & 92.86 & 75.30  & -\\
    SSL-HSIC \cite{li2021self} &  91.95  & 89.91 & 67.22  & 57.01 & 92.06  & 88.87 & 51.42  & 36.03 & 74.77  & 92.56 & 72.13  & 90.33\\
    VICReg \cite{bardesVICRegVarianceInvarianceCovarianceRegularization2022} &  91.08  & 88.93 & 66.91  & 56.47 & 91.11  & 86.24 & 50.17  & 34.24 & 74.88  & 92.84 & -  & -\\
    \midrule
    SimCLR + SparseHead &  \textbf{92.75}  & \textbf{90.32} & \textbf{67.55}  & 56.84 & 92.24  & 88.49 & 51.42  & 36.16 &  72.27 & 91.24 & 71.44  & 90.82\\
    BYOL + SparseHead &  92.45  & 90.05 & 67.52  & 56.86 & \textbf{92.60}  &  \textbf{88.63}  & \textbf{52.16} & \textbf{36.74}  & \textbf{75.94}& \textbf{92.91} & \textbf{75.52}  & \textbf{92.76} \\
    Barlow Twins + SparseHead &  91.28  & 89.87 & 66.91  & 56.47 & 91.45  & 88.46 & 51.49  & 35.26 &  75.75  & 92.95 & 75.32  & 92.28\\
    \bottomrule
    \end{tabular}}
\label{tab:result1}
\end{table}

\begin{minipage}{\textwidth}
\vspace{-0.1cm}
\begin{minipage}[t]{0.48\textwidth}
\makeatletter\def\@captype{table}
\caption{Semi-supervised classification. We finetune the pre-trained model using 1\% and 10\% training samples of ImageNet following \cite{zbontarBarlowTwinsSelfSupervised2021}.}
\vspace{0.1cm}
\resizebox{\linewidth}{!}{
\begin{tabular}{lccccc}
		\toprule
		\multirow{2}*{Methods} & \multirow{2}*{Epochs}  & \multicolumn{2}{c}{1\%} & \multicolumn{2}{c}{10\%} \\\cline{3-6}
		~ & ~ & top-1 & top-5 & top-1 & top-5 \\
		\midrule
		SimCLR & 1000 &  48.3  & 75.5 & 65.6  & 87.8\\
		BYOL & 1000 & 53.2  & 78.4 & 68.8 & 89.1\\
		SwAV & 1000 & 53.9  & 78.5 & 70.2  & 89.9\\
		BarlowTwins & 1000 & 54.9  & 79.4 & 69.6  & 89.1\\
		\midrule
		SimCLR + SparseHead & 1000 & 50.1 &  76.4 & 67.5  & 88.7\\
		BYOL + SparseHead & 1000 & 54.3  & 78.5 & 69.4  & 89.5\\
		Barlow Twins + SparseHead & 1000 & \textbf{56.2}  & \textbf{79.5} & \textbf{70.6} & \textbf{89.6}\\
		\bottomrule
	\end{tabular}}
\label{tab:semi-supervised}
\end{minipage}
\hspace{.1in}
\begin{minipage}[t]{0.48\textwidth}
\makeatletter\def\@captype{table}
\caption{The results of transfer learning on object detection and instance segmentation with C4-backbone as the feature extractor.}
\vspace{0.1cm}
\resizebox{\linewidth}{!}{
\begin{tabular}{lcccccc}
		\toprule
		\multirow{2}*{Methods} & \multicolumn{3}{c}{Object Det.} & \multicolumn{3}{c}{Instance Seg.} \\
		\cline{2-4}
		\cline{5-7}
		~ & AP & AP$_{50}$ & AP$_{75}$ & AP & AP$_{50}$ & AP$_{75}$ \\
		\midrule
		SimCLR & 37.9 &  57.7  & 40.9 & 33.2  & 54.6 & 35.3 \\
		SwAV & 37.6 &  57.6  & 40.2 & 33.0  & 54.2 & 35.1 \\
		BYOL & 37.9 &  57.8  & 40.9 & 33.1  & 54.3 & 35.0 \\
            SimSiam & 37.9 &  57.5  & 40.9 & 33.3  & 54.2 & 35.2 \\
		BarlowTwins & 39.2 & 59.0  & 42.5 & 34.2  & 56.0 & 36.5 \\
		\midrule
		SimCLR + SparseHead & 37.7 & 58.5 & 42.1 & 34.7 & 55.4 & 36.2\\
		BYOL + SparseHead & 39.5 & \textbf{59.8}  & 43.4 & \textbf{35.8} & 56.2 & \textbf{37.2}\\
		Barlow Twins + SparseHead & \textbf{39.9} & 59.6 & \textbf{43.8} & 35.4 & \textbf{56.4} & 36.9\\
		\bottomrule
	\end{tabular}}
\label{tab:transfer}
\end{minipage}
\end{minipage}

\section{Experiments}
\label{sec:Exp}

\subsection{Experimental Setup}
\textbf{Datasets}. We conducted experiments on the following datasets: CIFAR-10, CIFAR-100\cite{krizhevsky2009learning}, STL-10\cite{coatesAnalysisSingleLayerNetworks2011}, Tiny ImageNet\cite{leTinyImagenetVisual2015}, ImageNet 100\cite{tianContrastiveMultiviewCoding2020}, ImageNet\cite{dengImageNetLargescaleHierarchical2009} and COCO\cite{lin2014microsoft}. 

\textbf{Implementation Details}. In the linear evaluation task, we employed the ResNet-18 network on four datasets: CIFAR-10, CIFAR-100, STL-10, and Tiny ImageNet. For ImageNet and ImageNet100, we utilized the Resnet-50 network. The encoder was trained for 1000 epochs on CIFAR-10 and CIFAR-100, using a learning rate of $3\times10^{-3}$. On STL-10, we trained the encoder for 2000 epochs with a learning rate of $2\times10^{-3}$, while on Tiny ImageNet, we trained it for 1000 epochs with a learning rate of $2\times10^{-3}$. Regarding ImageNet and ImageNet100, we followed the settings outlined in \cite{chenSimpleFrameworkContrastive2020,grillBootstrapYourOwn2020,zbontarBarlowTwinsSelfSupervised2021} and trained for 1000 epochs. The Adam optimizer \cite{kingmaAdamMethodStochastic2014} was employed with a weight decay of $1\times10^{-6}$. All models utilized a projection head, consisting of an MLP with two fully connected layers, batch normalization, and ReLU activation function. Notably, in SimCLR, BYOL, and Barlow-Twins, we included the SparseHead regularization term within the projection head. The temperature parameter in Equation \ref{eq:infonce} was set to $\tau=0.5$, and the sparsity regularization parameter in Equation \ref{eq:problem} $\lambda=1\times10^{-4}$.

\textbf{Linear evaluation}. We test the generalization performance of the learned feature extractor following the same procedure as Equation \ref{eq:downstream} by freezing the $f_\theta$ and train a classifier $W$ on top of it. Additionally, for datasets that are small to medium-sized, we also evaluate the accuracy of a k-nearest neightbors classifier with $k=5$. Table \ref{tab:result1} presents the results on small, medium, and large-sized datasets. We incorporate the SparseHead into three models, namely SimCLR+SparseHead, BYOL+SparseHead, and Barlow Twins+SparseHead. From Table \ref{tab:result1}, it is evident that the performance of our proposed methods surpasses that of the baseline. Notably, the proposed methods exhibits the best results by enhancing the linear evaluation accuracy by more than 1\%. Specifically, SimCLR+SparseHead boosts the top 1 validation accuracy of SimCLR in ImageNet by 2.12\%, while BYOL+SparseHead enhances BYOL by 1.05\%.

\textbf{Semi-supervised learning}. We also assess the fine-tuning performance of the self-supervised pre-trained ResNet-50 model on a classification task with a subset of ImageNet's training set using labeled information. We adopt the semi-supervised approach introduced in \cite{kornblith2019better, chenSimpleFrameworkContrastive2020, zhai2019large, henaff2020data}, and employ the same fixed splits of 1\% and 10\% of labeled ImageNet training data, respectively. The test set's top 1 and top 5 accuracies are reported, and the models are trained for 1000 epochs, as demonstrated in Table \ref{tab:semi-supervised}. When only 1\% of the labels are available, the Barlow Twins+SparseHead model attains a top 1/top 5 accuracy of 56.2\%/79.5\%, improving the accuracy by 1.3\%/0.1\% points compared to the 200 pre-training epochs setting. The semi-supervised classification results suggest that SparseHead learns superior representations for image-level prediction.

\textbf{Transfer learning}. We evaluate the performance of SparseHead on the COCO dataset \cite{lin2014microsoft} using the instance segmentation and object detection tasks. Following the common transfer learning setting employed by existing methods \cite{zbontarBarlowTwinsSelfSupervised2021, grillBootstrapYourOwn2020}, we adopt Mask R-CNN \cite{he2017mask} with a $1\times$ schedule and the same backbone as Faster R-CNN. The results, presented in Table \ref{tab:transfer}, demonstrate that the proposed method significantly enhances the performance of downstream tasks.

\subsection{Ablation Study}
\textbf{The effective representations.} To demonstrate the improvement of generalization ability and solve the dimension collapse problem by the sparse projection head, we trained a feature extractors ResNet-50 on the STL-10 dataset. The representation dimension of ResNet-50 is $d=2048$. We obtained the eigenvalue spectra on the test set in the same way as the motivation experiment in Sec. \ref{sec:motivation}. The eigenvalue spectra of the representation vectors and embedding vectors with the sparse constraint and without the sparse constraint are shown in the following figure.
\begin{figure}[h]
    \centering
     \begin{subfigure}{0.45\textwidth}
         \centering
         \includegraphics[width=\textwidth]{fig4.eps}
         \caption{Without SparseHead}
         \label{fig:2.1}
     \end{subfigure}
     \begin{subfigure}{0.45\textwidth}
         \centering
         \includegraphics[width=\textwidth]{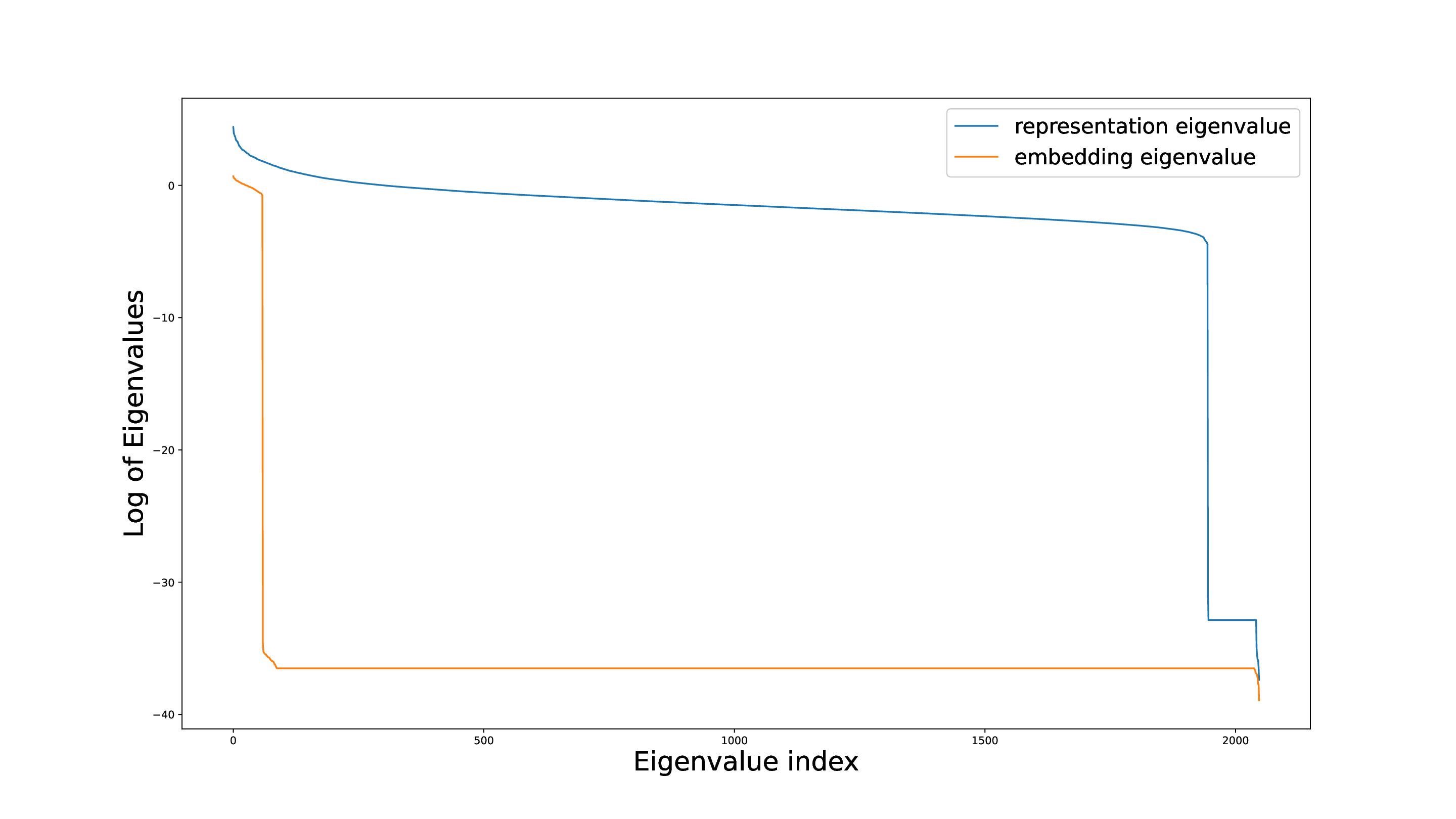}
         \caption{With SparseHead}
         \label{fig:2.2}
     \end{subfigure}
        \caption{The eigenvalue spectra of the representation vectors and embedding vectors with and without SparseHead}
        \label{fig:comparison}
\end{figure}

Figure \ref{fig:comparison} shows that when the higher dimensional feature space, the issue of dimension collapse persists even when a projection head is employed. However, by controlling the sparsity of the projection head, we observe that the effective dimension of the embedding decreases, while the effective dimension of the representation increases. This observation confirms the correctness of our previous conclusion.

\section{Conclusion}
In this work, we perform empirical studies to propose the key assumption that only a subset of features are used when minimizing the contrastive loss of a batch of samples. Then, we theoretically showed that constraining the sparsity of projection head can improve the genralization and discriminative of the learned feature extractor network. Inspired by our theory, we proposed SparseHead, a regulator that constrain the sparsity of projection head and can be seamlessly integrated with any self-supervised learning (SSL) approaches. Finally, by a series of experiments, we showed that our proposed SparseHead regulator significantly improve the performance of existing self-supervised method on linear evaluation and Semi-Supervised tasks.

\textbf{Limitations and Broader Impact}. The regularization hyperparameter $\lambda$ is pre-defined in this paper, improvements might be obtained by learning this parameter in an adaptive way. Another limitation is that our experiments are conducted only on ResNet architectures. Future research might incorporate architectural improvements, and demonstrate the advantage of the proposed SparseHead.







{\small
\bibliography{bibliography}
\bibliographystyle{plain}
}


\end{document}